\documentclass[10pt,twocolumn,letterpaper]{article}

\usepackage{cvpr} % uncomment this for the final submission
\usepackage{times}
\usepackage{epsfig}
\usepackage{graphicx}
\usepackage{amsmath}
\usepackage{amssymb}

\usepackage{booktabs}
\usepackage{lipsum}
\usepackage{multirow}
\usepackage{bm}
\usepackage{lineno}
\usepackage[table]{xcolor} 

\begin{document}

%%%%%%%%% TITLE
\title{A Controllable 3D Deepfake Generation Framework with Gaussian Splatting}

%author list
% \author{Wending Liu\\
% The University of Tokyo\\
% National Institute of Informatics\\
% Tokyo, Japan\\
% {\tt\small lwd@nii.ac.jp}
% \and
% Siyun Liang\\
% \and
% Huy H. Nguyen\\
% \and
% Isao Echizen\\
% The University of Tokyo\\
% National Institute of Informatics\\
% Tokyo, Japan\\
% {\tt\small iechizen@nii.ac.jp}
% }

\author{
    Wending Liu$^{1,2}$ \quad
    Siyun Liang$^{3}$ \quad
    Huy H. Nguyen$^{2}$\quad
    Isao Echizen$^{1,2}$ \vspace{0.4em} \\
    {\normalsize $^1$The University of Tokyo} \quad
    {\normalsize $^2$National Institute of Informatics} \quad
    {\normalsize $^3$Technical University of Munich} \\
}

%%%%%%%%% TEASER
\twocolumn[{%
    \renewcommand\twocolumn[1][]{#1}%
    \maketitle
    \thispagestyle{empty}
    \begin{center}
        \includegraphics[width=0.95\linewidth]{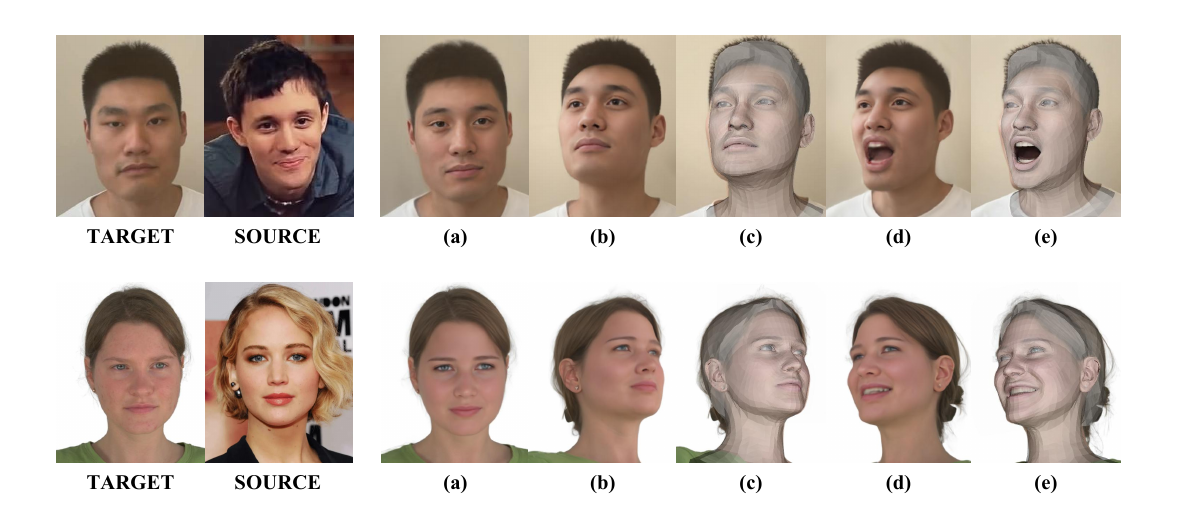}
        \captionof{figure}{Our method generates high-quality 3D deepfakes with animatable head models and spatially aligned background scenes. Given a source input, our method achieves photorealistic face swapping while preserving the target’s pose and expression, as well as the surrounding background (a). Leveraging a parametric head model and explicit camera modeling, we enable flexible control over viewpoints and expressions (b, d). As shown in (c) and (e), our method reconstructs complete 3D geometry, including detailed depth information.}
        % Compared to prior 2D methods, our approach ensures more realistic face swapping, better preservation of pose and expression, real-time rendering, and improved depth consistency for defense against depth-based deepfake detection.
        \label{fig:teaser}
    \end{center}
}]

\maketitle
\thispagestyle{empty}

%%%%%%%%% ABSTRACT
\begin{abstract}
   We propose a novel 3D deepfake generation framework based on 3D Gaussian Splatting that enables realistic, identity-preserving face swapping and reenactment in a fully controllable 3D space. Compared to conventional 2D deepfake approaches that suffer from geometric inconsistencies and limited generalization to novel view, our method combines a parametric head model with dynamic Gaussian representations to support multi-view consistent rendering, precise expression control, and seamless background integration. To address editing challenges in point-based representations, we explicitly separate the head and background Gaussians and use pre-trained 2D guidance to optimize the facial region across views. We further introduce a repair module to enhance visual consistency under extreme poses and expressions. Experiments on NeRSemble and additional evaluation videos demonstrate that our method achieves comparable performance to state-of-the-art 2D approaches in identity preservation, as well as pose and expression consistency, while significantly outperforming them in multi-view rendering quality and 3D consistency. Our approach bridges the gap between 3D modeling and deepfake synthesis, enabling new directions for scene-aware, controllable, and immersive visual forgeries, revealing the threat that emerging 3D Gaussian Splatting technique could be used for manipulation attacks.
\end{abstract}

%%%%%%%%% BODY TEXT
\section{Introduction}
\label{sec:intro}

Deepfake technology has rapidly advanced in recent years, driven by progress in deep learning~\cite{krizhevsky2017imagenet} and generative models~\cite{goodfellow2014generative, kingma2013vae, ho2020ddpm}, enabling the synthesis of highly realistic faces, voices, and expressions~\cite{misirlis2023deepfake}. While 2D-based approaches have demonstrated impressive visual quality in face swapping, expression manipulation, and voice-driven animation~\cite{pei2024deepfake}, they still suffer from fundamental limitations.

Specifically, 2D methods often exhibit perceptible artifacts under large pose or expression variations, compromising visual realism. Moreover, with the growing prevalence of depth-based detection techniques, 2D-generated content becomes increasingly vulnerable to forgery detection. In scenarios that require long video generation, multi-view consistency, or immersive applications, the lack of 3D structure leads to poor geometric consistency and limited controllability, posing severe constraints on practical deployment.

Although recent research has made notable progress in 3D face reconstruction~\cite{retsinas20243d}, relatively little attention has been paid to 3D deepfake generation and controllable face swapping~\cite{felouat20253ddgd}. In addition, most existing methods neglect background modeling, making it difficult to synthesize coherent content across multiple viewpoints. The integration of animatable 3D head models with view-consistent 3D backgrounds remains largely unexplored, significantly limiting the realism and applicability of current systems.

To address these challenges, we propose a complete 3D deepfake generation framework that combines 3D Gaussian Splatting~\cite{kerbl20233dgs} with the FLAME head model~\cite{li2017flame} to jointly model both the head and background. As shown in Figure~\ref{fig:teaser}, our framework is highly controllable, delivering high-quality visual results that remain robust even under extreme poses and expressions. The generated outputs possess accurate 3D geometry and consistent depth, making the generated results more resilient against detection methods based on depth cues and multi-view geometric consistency. By incorporating 3D backgrounds, we fully exploit the advantages of head modeling, while the efficient rendering process ensures applicability in real-time scenarios such as live video conferencing.

We validate the effectiveness of our method on the NeRSemble dataset~\cite{kirschstein2023nersemble} and through physical evaluations involving real-world scenes and camera settings.

Our main contributions are summarized as follows:
\begin{enumerate}
    \item We propose a novel 3D deepfake generation framework based on Gaussian Splatting, enabling realistic face swapping and reenactment in 3D space.
    \item We introduce precise facial motion control using FLAME parameters, allowing for fine-grained manipulation of expressions and poses across various views.
    \item We design a 3D alignment strategy to synchronize animatable head models with reconstructed backgrounds using optimized camera parameters.
    \item We propose a data preparation pipeline that enables users to easily construct input data compatible with our framework, supporting controllable 3D deepfake synthesis and facilitating future research in this area.
\end{enumerate}

\section{Related Work}
\label{sec:related_work}

\paragraph{2D Deepfake Generation.} 2D deepfake methods have achieved remarkable progress in recent years~\cite{sharma2024systematic}, enabling high-quality face swapping, reenactment, and manipulation from single-view images or videos. Notable works include StyleGAN~\cite{karras2020stylegan2} and other GAN-based methods~\cite{sun2024activefakedeepfakecamouflage, choi2024latentswapefficientlatentcode, 10.1145/3676165, 10536627, li20233dswap}, which provides photorealistic facial synthesis via latent space manipulation, and FaceShifter~\cite{li2019faceshifter}, which improves identity preservation in face swapping through two-stage alignment. Other methods such as SimSwap~\cite{chen2020simswap} and FaceForensics++~\cite{rossler2019faceforensics++} focus on expression consistency and face forgery detection benchmarks. In addition, recent diffusion-based approaches~\cite{wang2025dynamicfacehighqualityconsistentface, baliah2024realisticefficientfaceswapping, zhao2023diffswap} have demonstrated strong capabilities in generating realistic and temporally consistent DeepFakes, further advancing the state of the art. Despite these successes, 2D-based methods are fundamentally limited by their lack of geometric awareness. They often struggle under large pose variations, exaggerated expressions, or inconsistent lighting, as they operate without 3D structure. Moreover, they provide limited control over facial motion and typically lack multi-view consistency, making them unsuitable for immersive applications such as virtual reality.

\paragraph{3D Face Reconstruction and Animation.} To overcome the spatial limitations of 2D approaches, 3D face modeling has been extensively studied. Traditional methods rely on parametric models such as 3D Morphable Models (3DMMs)~\cite{blanz2023morphable} and FLAME~\cite{li2017flame}, which represent facial geometry via low-dimensional shape, expression, and pose parameters. These models are compact and interpretable, making them suitable for expression editing and animation. However, their limited resolution and inability to capture fine-grained details (e.g., skin texture, wrinkles, and hair) reduce realism in downstream applications~\cite{liang20253dmasterface}. Learning-based methods improve this. DECA~\cite{feng2021deca} adds a detail map for high-frequency features, and MICA~\cite{zielonka2022mica} uses contrastive learning to refine identity embeddings. However, these models are designed for analysis and animation, not deepfake generation. Furthermore, most of them reconstruct only the head region without considering background or camera context, limiting their realism in scene-aware synthesis.

\begin{figure*}
    \centering 
    \includegraphics[width=\linewidth]{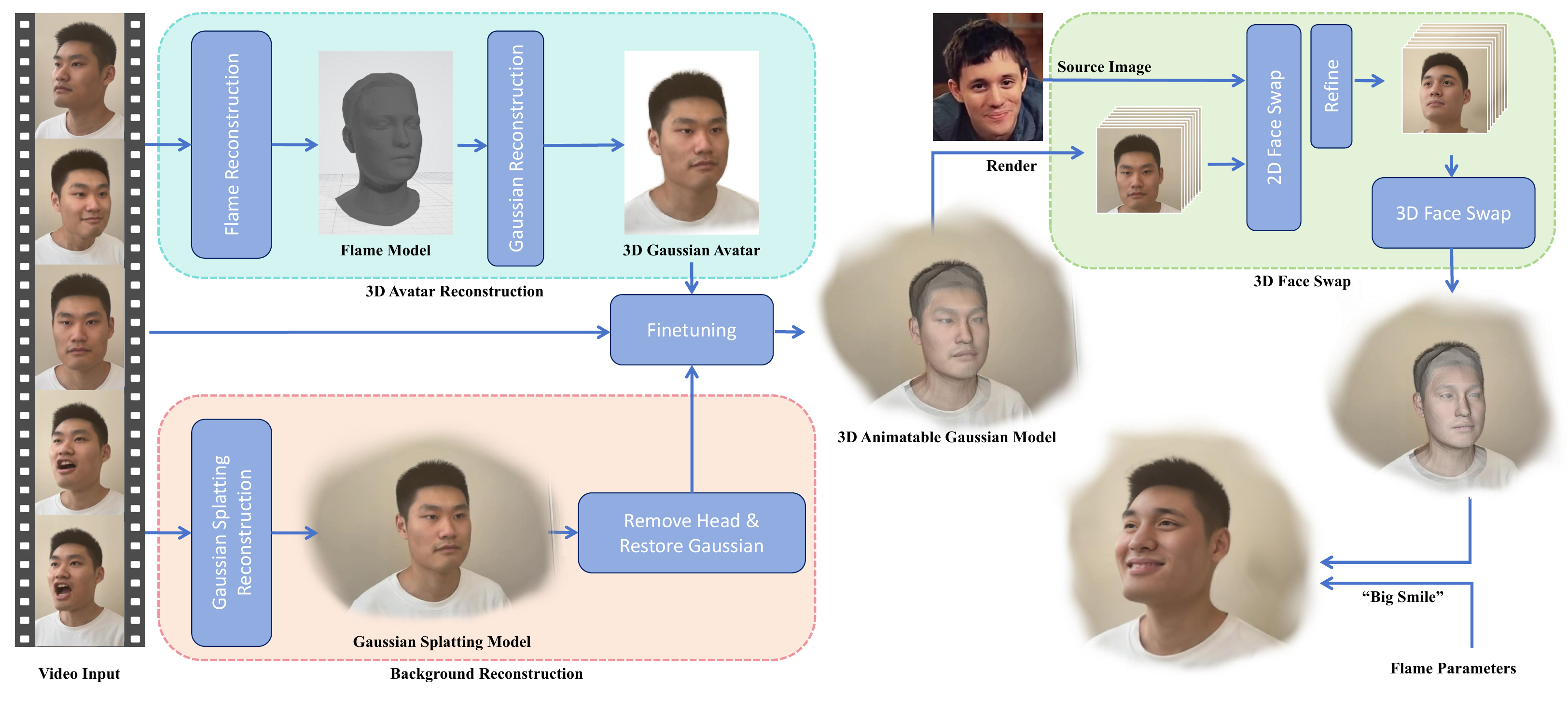} 
    \caption{\textbf{Pipeline Overview.} Our pipeline consists of three main components: FLAME- and Gaussian Splatting-based 3D head reconstruction, 3D face swapping guided by 2D baseline supervision, and 3D background reconstruction with alignment. The resulting model supports expression and motion control via FLAME parameters, enabling high-quality and controllable 3D DeepFake generation.}
    \label{fig:overview}
\end{figure*}

\paragraph{3D Representations for Scene and Face Modeling.} The underlying 3D representation significantly impacts the visual fidelity, efficiency, and controllability of generative systems~\cite{ze20243d}. Mesh-based representations~\cite{pabitha2024dense, zhang2024t, weng2024pivotmesh}, as used in traditional modeling pipelines, provide explicit geometry and are well-suited for animation, but suffer from topological complexity and poor scalability for fine details like hair or dynamic expressions. In contrast, implicit representations such as Neural Radiance Fields (NeRF)~\cite{mildenhall2021nerf} encode volumetric scenes through neural networks, allowing photorealistic view synthesis. Though expressive, NeRF-based models require dense multi-view input, are computationally expensive, and lack explicit control over expression or identity~\cite{wang20243d}. Efforts to apply NeRF to human faces~\cite{hong2022headnerf, 10.1145/3550454.3555501, gafni2020dynamicneuralradiancefields} face challenges in temporal consistency and animation control. Recently, 3D Gaussian Splatting~\cite{kerbl20233dgs} has emerged as a more efficient and expressive alternative. It represents scenes with spatially distributed Gaussians, enabling real-time rendering and preserving geometric detail. It also integrates well with standard 3D tools and GPU pipelines, making it ideal for dynamic head modeling and background alignment in 3D deepfake generation.

% \begin{figure*}
%     \centering 
%     \includegraphics[width=\linewidth]{fig/pipeline.pdf} 
%     \caption{\textbf{Pipeline Overview.} Our pipeline consists of three main components: FLAME- and Gaussian Splatting-based 3D head reconstruction, 3D face swapping guided by 2D baseline supervision, and 3D background reconstruction with alignment. The resulting model supports expression and motion control via FLAME parameters, enabling high-quality and controllable 3D DeepFake generation.}
%     \label{fig:overview}
% \end{figure*}

\section{Method}
\label{sec:method}

In this section, we present a novel 3D DeepFake generation framework based on 3D Gaussian Splatting, which enables high-fidelity and highly controllable facial reenactment and identity replacement across arbitrary viewpoints.

An overview of our pipeline is illustrated in Figure~\ref{fig:overview}.

\subsection{FLAME-Based Head Modeling with 3D Gaussian Splatting}

We adopt a novel approach that integrates the FLAME model with 3D Gaussian Splatting to construct an animatable 3D head representation, combining strong expressiveness with efficient rendering capabilities. FLAME is a parametric 3D face model that compactly represents identity, expression, and pose using a small set of parameters. It defines facial geometry using three types of parameters: shape $\boldsymbol{\beta}$, expression $\boldsymbol{\psi}$, and pose $\boldsymbol{\theta}$. The canonical mesh deformation is defined as:

\begin{equation}
    T_{\mathcal{P}}(\boldsymbol{\beta}, \boldsymbol{\theta}, \boldsymbol{\psi}) = \overline{T} + B_S(\boldsymbol{\beta}; \mathcal{S}) + B_P(\boldsymbol{\theta}; \mathcal{P}) + B_E(\boldsymbol{\psi}; \mathcal{E}),
\end{equation}

where $\overline{T}$ denotes the mean template mesh and $B_S$, $B_P$ and $B_E$ represent weighted deformations generated by linear bases for shape, pose, and expression, respectively. The final posed mesh is obtained by linear blend skinning (LBS~\cite{SMPL:2015}):

\begin{equation}
    M(\boldsymbol{\beta}, \boldsymbol{\theta}, \boldsymbol{\psi}) = W(T_{\mathcal{P}}(\boldsymbol{\beta}, \boldsymbol{\theta}, \boldsymbol{\psi}), J(\boldsymbol{\beta}), \boldsymbol{\theta}, \mathbf{W}),
\end{equation}

where $J(\boldsymbol{\beta})$ specifies the joint positions and $\mathcal{W}$ denotes the weights of the skin of the vertices.

To enhance the rendering quality and enable real-time synthesis, we adopt the 3D Gaussian Splatting representation. Each Gaussian is parameterized by a center position $\boldsymbol{\mu}_i$, a covariance matrix $\bm{\Sigma}_i$, an opacity $\sigma_i$, and appearance attributes (e.g. color). Following Kerbl et al.~\cite{jena2023splatarmor}, the covariance matrix $\boldsymbol{\Sigma}_i$ is constructed from a diagonal scaling matrix $\bm{S} \in \mathbb{R}^{3 \times 3}$, which is constructed from a scaling vector $\boldsymbol{s}_i \in \mathbb{R}^3$, and a rotation matrix $\boldsymbol{R}_i \in \mathbb{R}^{3 \times 3}$ as:

\begin{equation}
    \bm{\Sigma} = \bm{R} \bm{S} \bm{S}^T \bm{R}^T.
\end{equation}

In practice, we store each Gaussian using a position vector $\boldsymbol{\mu}_i$, a scaling vector $\boldsymbol{s}_i$, and a quaternion $\boldsymbol{q}_i \in \mathbb{R}^4$. The quaternion $\boldsymbol{q}_i$ is converted to a rotation matrix $\boldsymbol{R}_i$ for the computation of covariance during rendering.

For rendering, the color C of a pixel is computed by
blending all 3D Gaussians overlapping the pixel:

\begin{equation}
    \mathbf{C} = \sum_{i=1}^{n} c_i \alpha_i' \prod_{j=1}^{i-1} (1 - \alpha_j'),
\end{equation}

where $c_i$ is the color of each point, modeled by 3-degree spherical harmonics. The weight of blending $\alpha_i'$ is given by evaluating the 2D projection of the 3D Gaussian, multiplied by the opacity per point $\alpha_i$. The Gaussian splats are sorted by depth before blending to respect visibility order.

The set of Gaussians is projected onto the image plane and rendered using a forward rasterization process involving depth sorting and soft alpha blending, enabling real-time multiview image synthesis.

To animate Gaussians with FLAME-driven motion, we follow the GaussianAvatars approach~\cite{qian2024gaussianavatars} and bind each Gaussian to a corresponding triangle on the FLAME mesh. Each triangle hosts a Gaussian centered at its barycenter, and the local-to-global transformation of a Gaussian is defined as:
\begin{align*}
\bm{r}' &= \mathbf{R'}\bm{r}, \\
\bm{\mu}' &= k\mathbf{R'}\bm{\mu} + \mathbf{T'}, \\
\bm{s}' &= k\bm{s},
\end{align*}

where $\mathbf{R'}$ is the local rotation matrix constructed from the triangle orientation, $\mathbf{T'}$ is the barycenter of the triangle, and $k$ is a global scale factor. Through this binding, Gaussians are transformed consistently with the mesh deformation, allowing for pose and expression-driven animation.

\subsection{3D Face Swapping with 2D Supervision}
Building upon the FLAME-based animatable 3D head representation, we propose a 3D face swapping pipeline guided by 2D supervision. While existing 2D methods~\cite{chen2020simswap, zhao2023diffswap, li20233dswap, gao2021infoswap} produce visually plausible results, they suffer from (1) a lack of 3D geometric consistency, leading to distortions under large poses or multi-view settings, and (2) vulnerability to depth-based forgery detection due to the absence of spatial structure.

To address these issues, we embed prior knowledge from 2D face swapping models into a 3D Gaussian representation. Specifically, we use a pretrained 2D model~\cite{chen2020simswap} to generate multi-view swapped images from a given source identity. These serve as supervision to guide the transformation of Gaussian attributes (e.g., color and opacity), while the FLAME-Gaussian structure preserves 3D consistency and provides realistic depth cues.

To ensure identity consistency across views, we employ multi-view supervision using pixel-wise losses such as L1, LPIPS~\cite{zhang2018unreasonableeffectivenessdeepfeatures}, and perceptual loss to align 3D renderings with 2D swapped images.

To avoid affecting the background or irrelevant regions, we apply localized identity optimization by restricting updates to facial Gaussians using segmentation masks, along with regularization.

Finally, to enhance visual quality under challenging expressions or occlusions, we incorporate an image-space refinement module based on CodeFormer~\cite{zhou2022robustblindfacerestoration}.
This Refine Unit addresses a key issue: 2D face swapping models may produce artifacts under extreme poses, resulting in noisy supervision that can impair 3D training. By refining these outputs before optimizing the 3D Gaussians, our module mitigates the impact of low-quality frames—particularly around sensitive regions like the mouth and eyes. It leverages global semantics and local detail restoration to provide more reliable targets, improving training stability and finer appearance modeling.

By integrating strong 2D priors into a structured 3D framework, our method achieves high-fidelity identity preservation and geometric consistency, improving realism and robustness under pose variation, expression changes, and depth-based detection.

\subsection{3D Background Reconstruction and Alignment}

While our method supports controllable 3D facial expression modeling and identity swapping, the absence of a background limits its applicability for generating realistic forged images or videos. To fully exploit the advantages of the 3D head model, it is essential to reconstruct a 3D background that is both geometrically aligned and visually consistent with the foreground.

Unlike traditional mesh-based 3D representations, 3D Gaussian Splatting encodes lighting information within each Gaussian using spherical harmonics~\cite{kerbl20233dgs}. As a result, lighting cannot be globally controlled through environmental light inputs. Consequently, reconstructing a background under the same lighting conditions as the 3D head model is critical. To this end, we propose a joint reconstruction and alignment strategy that enables coherent rendering of the head and background in the same 3D scene.

\subsubsection{Real-Scale Background Reconstruction}
We reconstruct the background using 3D Gaussian Splatting and ensure that it is scale-consistent with the FLAME-based head model. Since the FLAME model operates in real-world metric units (meters), but its parameters are typically regressed without access to actual camera intrinsics, its resulting scale may differ from COLMAP~\cite{schoenberger2016sfm}-based background reconstructions.

To resolve this discrepancy, we introduce a low-cost data acquisition strategy for obtaining real-scale camera poses. Although high-precision methods such as structured light or motion tracking systems can yield accurate poses, they are often expensive and difficult to deploy. Instead, we leverage the fact that modern smartphones (e.g., iPhone) are equipped with IMU sensors and support real-time pose tracking via development frameworks such as ARKit. By recording multi-view videos using a smartphone and exporting the per-frame camera extrinsics, we obtain an image sequence with accurate metric scale.

These real-world camera parameters are then fed into COLMAP to reconstruct a sparse point cloud, which is subsequently converted into a 3D Gaussian representation for the background, ensuring spatial consistency with the head model.

To remove the person from the reconstructed background and avoid redundant overlap, we first apply a segmentation model~\cite{kirillov2023segment} to extract 2D human masks in each frame, which are then back-projected into the 3D Gaussian space to assign semantic labels to the Gaussians, following GaussianEditor~\cite{chen2024gaussianeditor}. By analyzing their frequency and visibility across views, we obtain an accurate 3D person mask. Gaussians labeled as “person” are removed, and the resulting holes are filled using a diffusion-guided inpainting strategy~\cite{podell2023sdxl}. We perform morphological dilation around the boundaries and supervise reconstruction with synthetic inpainted images to ensure visual and structural consistency.

Through this process, we obtain a clean, high-quality, and spatially aligned 3D Gaussian background that serves as a robust foundation for downstream face swapping and scene synthesis tasks.

\subsubsection{Alignment and Joint Rendering}

Once the background is reconstructed, it must be aligned with the 3D head model for joint rendering in a shared coordinate space. However, most image-driven FLAME fitting pipelines do not recover the absolute position of the head in the real-world coordinate system. Instead, the pose parameters $(\mathbf{R}, \mathbf{t})$ absorb view-dependent variations, localizing the head only in camera-relative space.

To enable photorealistic 3D composition, it is necessary to transform the head model from its local (camera-relative) coordinate system into the global world coordinate system consistent with the background. Let $\mathbf{x}$ denote the coordinates of an arbitrary point in the world coordinate system, the rigid transformation of the FLAME model is defined as:
\begin{equation}
    \mathbf{x}_{\text{flame}} = \mathbf{R}(\mathbf{x} - \mathbf{x}_{\text{root}}) + \mathbf{x}_{\text{root}} + \mathbf{t} = \mathbf{R} \mathbf{x} + \mathbf{t}',
\end{equation}

where $\mathbf{x}_{\text{root}}$ is the root joint of the FLAME model, and the adjusted translation is given by $\mathbf{t}' = \mathbf{t} + (\mathbf{I} - \mathbf{R}) \mathbf{x}_{\text{root}}$, where $\mathbf{I}$ denotes the identity matrix.

We define two camera matrices:
\begin{enumerate}
    \item $\mathbf{W2C}_1$: the virtual camera used for rendering the canonical FLAME head model;
    \item $\mathbf{W2C}_2$: the actual camera matrix obtained during background reconstruction.
\end{enumerate}

To ensure consistent projections between the head and the background, we solve for a rigid transformation $\mathbf{T}_c = (\mathbf{R}_c, \mathbf{t}_c)$ that maps the head model into the background’s world coordinate system, such that:

\begin{equation}
    \mathbf{W2C}_1 (\mathbf{R} \mathbf{x} + \mathbf{t}') = \mathbf{W2C}_2 (\mathbf{R}_c \mathbf{x} + \mathbf{t}_c).
\end{equation}

Solving this equation yields the desired alignment transformation $\mathbf{T}_c$, which accurately places the head model into the reconstructed 3D scene, enabling seamless fusion between the foreground and background.

Once spatial alignment is established, both the head and background Gaussians are jointly passed through the Gaussian Splatting rendering pipeline. Although foreground regions in the background were masked and inpainted during preprocessing, low-density residual artifacts may still remain near the boundaries. To mitigate erroneous blending at these regions, we assign a higher rendering priority to the head Gaussians during rasterization. Specifically, during depth sorting prior to alpha blending, head Gaussians are rendered before background Gaussians, ensuring that the head projection dominates in overlapping regions. This strategy effectively eliminates ghosting and preserves spatial continuity in the final composite.

\section{Experiments}
\label{sec:exp}

\subsection{Experimental Setup}

\begin{table*}[htpb]
\setlength{\tabcolsep}{4.5pt}
% \small
\centering
% \resizebox{\textwidth}{!}{
    \begin{tabular}{l|ccccc|ccccc}
    \toprule
    &\multicolumn{5}{c|}{NeRSemble~\cite{kirschstein2023nersemble}}  & \multicolumn{5}{c}{Physical Evaluation} \\
    Method & $\text{ID} \uparrow$ & $\text{ID}_{\text{2D}}\uparrow$ & $\text{ID}_{\text{3D}}\uparrow$ & Pose$\downarrow$ & Exp$\downarrow$ & $\text{ID} \uparrow$ & $\text{ID}_{\text{2D}} \uparrow$ & $\text{ID}_{\text{3D}} \uparrow$ & Pose$\downarrow$ & Exp$\downarrow$ \\
    \midrule
    SimSwap~\cite{chen2020simswap} & \cellcolor{red!20}0.7356 & \cellcolor{red!20}0.7749 & \cellcolor{yellow!20}0.6818 & \cellcolor{orange!20}9.7110 & \cellcolor{orange!20}4.6877 & \cellcolor{red!20}0.7119 & \cellcolor{red!20}0.8091 & 0.6956 & \cellcolor{red!20}8.5946 & \cellcolor{yellow!20}3.8364  \\
    InfoSwap~\cite{gao2021infoswap} & \cellcolor{yellow!20}0.5352 & \cellcolor{yellow!20}0.6464 & 0.6861 & 17.4641 & 6.7438 & \cellcolor{yellow!20}0.5687 & \cellcolor{yellow!20}0.6765 & \cellcolor{orange!20}0.7134 & 36.2198 & 6.3101   \\
    % MegaFS~\cite{zhu2021megafs} & - & - & - & - & - & - & - & - & - & - & - & -  \\
    DiffSwap~\cite{zhao2023diffswap} & 0.2636 & 0.4309 & \cellcolor{orange!20}0.6890 & \cellcolor{yellow!20}12.9707 & \cellcolor{yellow!20}5.1099 & 0.3046 & 0.4758 & \cellcolor{yellow!20}0.7132 & \cellcolor{yellow!20}10.7149 & \cellcolor{orange!20}3.6823   \\
    3DSwap~\cite{li20233dswap} & 0.1310 & 0.3306 & 0.6667 & 35.5688 & 6.4208 & 0.1877 & 0.3703 & 0.6987 & 31.9860 & 6.4441   \\
    \textbf{Ours} & \cellcolor{orange!20}0.6706 & \cellcolor{orange!20}0.7608 & \cellcolor{red!20}\textbf{0.7935} & \cellcolor{red!20}\textbf{9.0358} & \cellcolor{red!20}\textbf{4.5623} & \cellcolor{orange!20}0.6768 & \cellcolor{orange!20}0.7715 & \cellcolor{red!20}\textbf{0.8724} & \cellcolor{orange!20}9.5886 & \cellcolor{red!20}\textbf{3.6473}   \\
    \bottomrule
    \end{tabular}
% }
\caption{\textbf{Quantative Results}. Our method better preserves the target’s pose and expression, achieves strong 3D identity consistency, and maintains 2D identity similarity and consistency on par with state-of-the-art 2D baselines.}
\label{tab:quan_1}
\end{table*}

\begin{table}
% \resizebox{\linewidth}{!}{
\centering
\begin{tabular}{c|c|c}
\toprule
SimSwap~\cite{chen2020simswap} & 
InfoSwap~\cite{gao2021infoswap} & \textbf{ours} \\
\midrule
3.29 & 1.88 & \\
\cmidrule{1-2}
DiffSwap~\cite{zhao2023diffswap} & 3DSwap~\cite{li20233dswap} & \textbf{14.45} \\
\cmidrule{1-2}
0.43 & - & \\
\bottomrule
\end{tabular}
% }
\caption{\textbf{Quantitative comparison on FPS $\uparrow$} of cross-identity facial reenactment. Our method is 4x times faster than the conventional 2D methods.}
\label{tab:fps}
\end{table}

\paragraph{Implementation Details}
Our framework is implemented in Python 3.9 with PyTorch and CUDA 11.8, running on Ubuntu 20.04. It is memory-efficient and can run on mid-range GPUs such as RTX 2080. We use the Adam optimizer with a learning rate scheduler to stabilize training.

The background Gaussian field is trained for 30k iterations, the animatable head for 600k iterations following~\cite{qian2024gaussianavatars}, and the 3D FaceSwap module for 300k iterations. During FaceSwap training, we further reduce the learning rate by 10× to improve fine-tuning stability. To maintain structural consistency, we disable the opacity reset commonly used in Gaussian Splatting~\cite{kerbl20233dgs}.

% Our method is compatible with various GPUs and has been tested on NVIDIA RTX 4070, RTX 2080, and A100. 

\paragraph{Dataset} We evaluate our method on a subset of the NeRSemble dataset~\cite{kirschstein2023nersemble}, which contains multi-view video recordings of 7 subjects from 16 synchronized cameras. For each subject, we sample 200 frames and 8 views per frame, yielding 1,600 images per subject for running 2D face-swapping baselines. These images covers diverse head poses and expressions, posing signifiant challenges for 2D methods. For comparison, we render our swapped 3D Gaussian avatars into the corresponding sampled frames. 

As the NeRSemble dataset lacks background information, to evaluate our method in a more challenging setting, we capture monocular videos of 3 subjects with background scenes with our data preparation pipeline. For each video (captured at 30 fps, approximately 1 minute in length), the camera moves around the subject while the subject performs a variety of exaggerated facial expressions. All frames are resized to 960×720 for processing. We then sample 200 frames per subject for evaluation. Permissions for research usage were obtained from all participants.

\paragraph{Baseline and metrics} We compare our method against several state-of-the-art 2D face swapping approaches, including SimSwap~\cite{chen2020simswap}, InfoSwap~\cite{gao2021infoswap}, DiffSwap~\cite{zhao2023diffswap}, and 3DSwap~\cite{li20233dswap}. 
% Additionally, we include ImplicitDeepfake~\cite{stanishevskii2024deepfake}, a recent 3DGS-based method, as a representative 3D baseline. 

Following standard practice, we employ a diverse set of quantitative metrics, summarized in Table~\ref{tab:quan_1}. To evaluate \textbf{identity preservation}, we compute the cosine similarity between the face embeddings of the source and swapped targets using a pre-trained 2D face recognition network~\cite{deng2019arcface}. We assess \textbf{pose error} as the $\mathcal{L}_2$ distance between the estimated head poses~\cite{martyniuk2022dad} of the target and swapped faces. Specifically, head pose is represented using Euler angles (in degrees). Additionally, we estimate the expression embeddings of both the target and the swapped faces using the method of~\cite{martyniuk2022dad}, and compute the $\mathcal{L}_2$ distance between them to quantify the expression error. These metrics collectively measure how well the swapped face preserves the identity of the source while accurately transferring the pose and expression of the target.

To further evaluate the 3D-awareness of face swapping methods, we introduce an \textbf{ID consistency} metric. Using the multi-view NeRSemble dataset, we randomly sample 200 frames and, for each, select 8 random views to assess multi-view ID consistency. For the physical evaluation, we directly compute ID consistency across the 200 sampled frames from each video. We further extend this evaluation to the 3D domain by assessing \textbf{3D ID consistency} using a 3D face recognition model~\cite{mu2019led3d}, which takes depth images as input. Since most 2D baselines~\cite{chen2020simswap, gao2021infoswap, zhao2023diffswap, li20233dswap} do not provide 3D geometry, we estimate depth maps from their swapped 2D face images using DepthAnythingV2~\cite{depth_anything_v2}. For our method, we directly render the underlying flame model to get smooth and consistent depth maps.

\begin{figure*}
    \centering 
    \includegraphics[width=\linewidth]{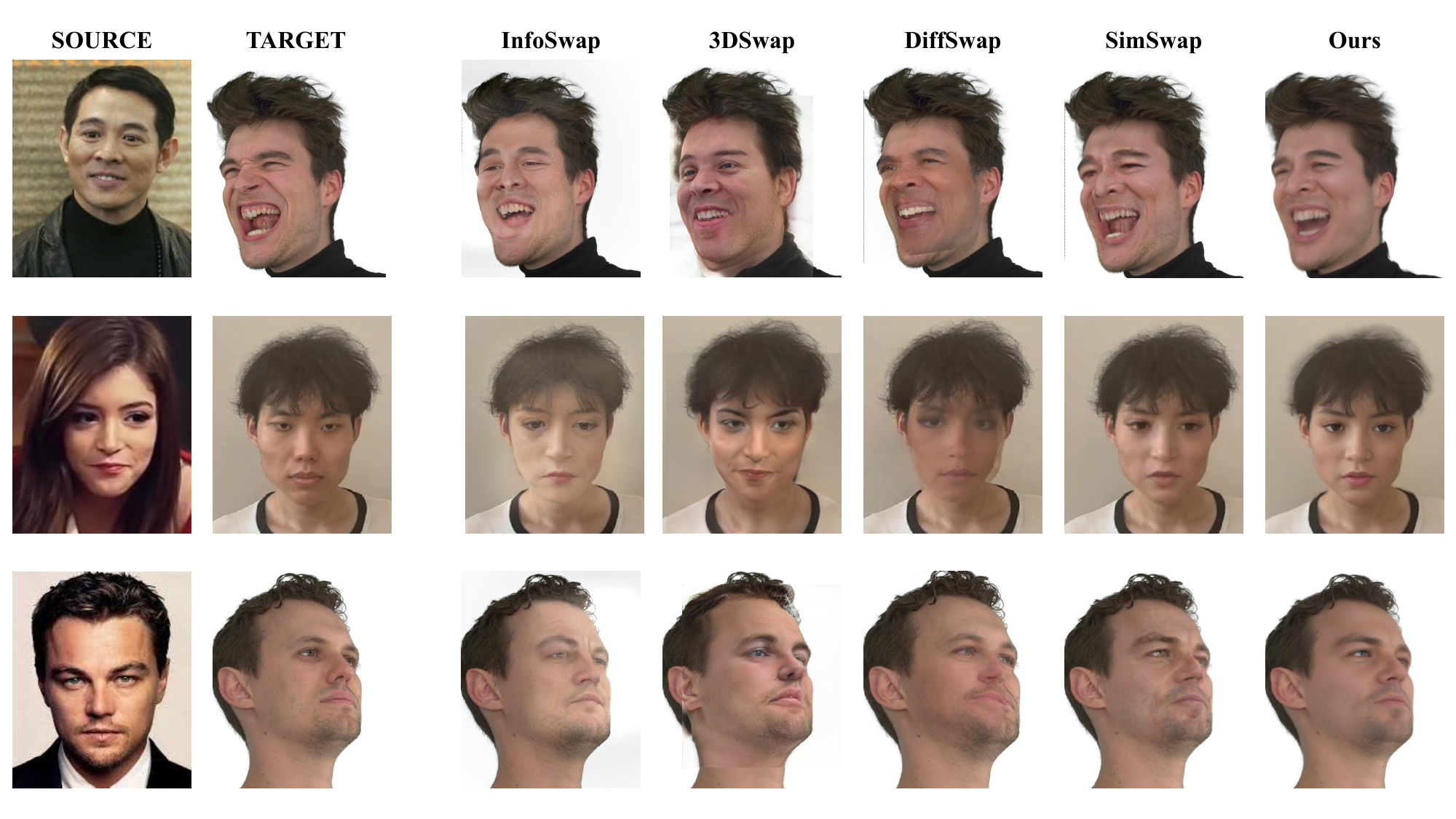} 
    \caption{\textbf{Comparison with 2D face swapping methods.} We compare our method with existing state-of-the-art 2D face swapping methods. The first row shows results under exaggerated expressions, the second row under neutral expressions, and the third row under challenging head poses. Our method achieves better identity preservation and visual consistency across all cases.}
    \label{fig:compare}
\end{figure*}

\subsection{Quantitative Comparison}
We report the quantitative comparison in Table~\ref{tab:quan_1}. Notably, we use the 2D deepfake generated from SimSwap~\cite{chen2020simswap} as supervision to fine-tune our 3D face swap. As shown in Table~\ref{tab:quan_1}, our method achieves lower pose and expression errors compared to 2D-based baselines. We attribute this improvement to the use of an explicit 3D face model and multi-view optimization, which enforce geometric and appearance consistency across different viewpoints. This multi-view consistency not only reduces artifacts under extreme head poses and facial expressions, but also contributes to more reliable depth cues, leading to stronger 3D ID consistency.

However, this consistency comes with a trade-off. By optimizing for global coherence across views, our method avoids overfitting to the source identity in any single frame. As a result, the 2D identity similarity is slightly lower than our 2D supervision~\cite{chen2020simswap}, which can tightly preserve identity in isolated frames but often lack cross-view consistency. Nonetheless, the decrease in 2D identity similarity is relatively minor, and our method still retains strong source identity preservation thanks to supervision from high-quality 2D deepfake outputs. Moreover, our framework is highly flexible and agnostic to the specific 2D method used for guidance. As a result, the 2D identity consistency of our system can be further improved by incorporating more advanced or specialized 2D face-swapping models as supervision.

\begin{figure}[htpb]
    \centering 
    \includegraphics[width=\linewidth]{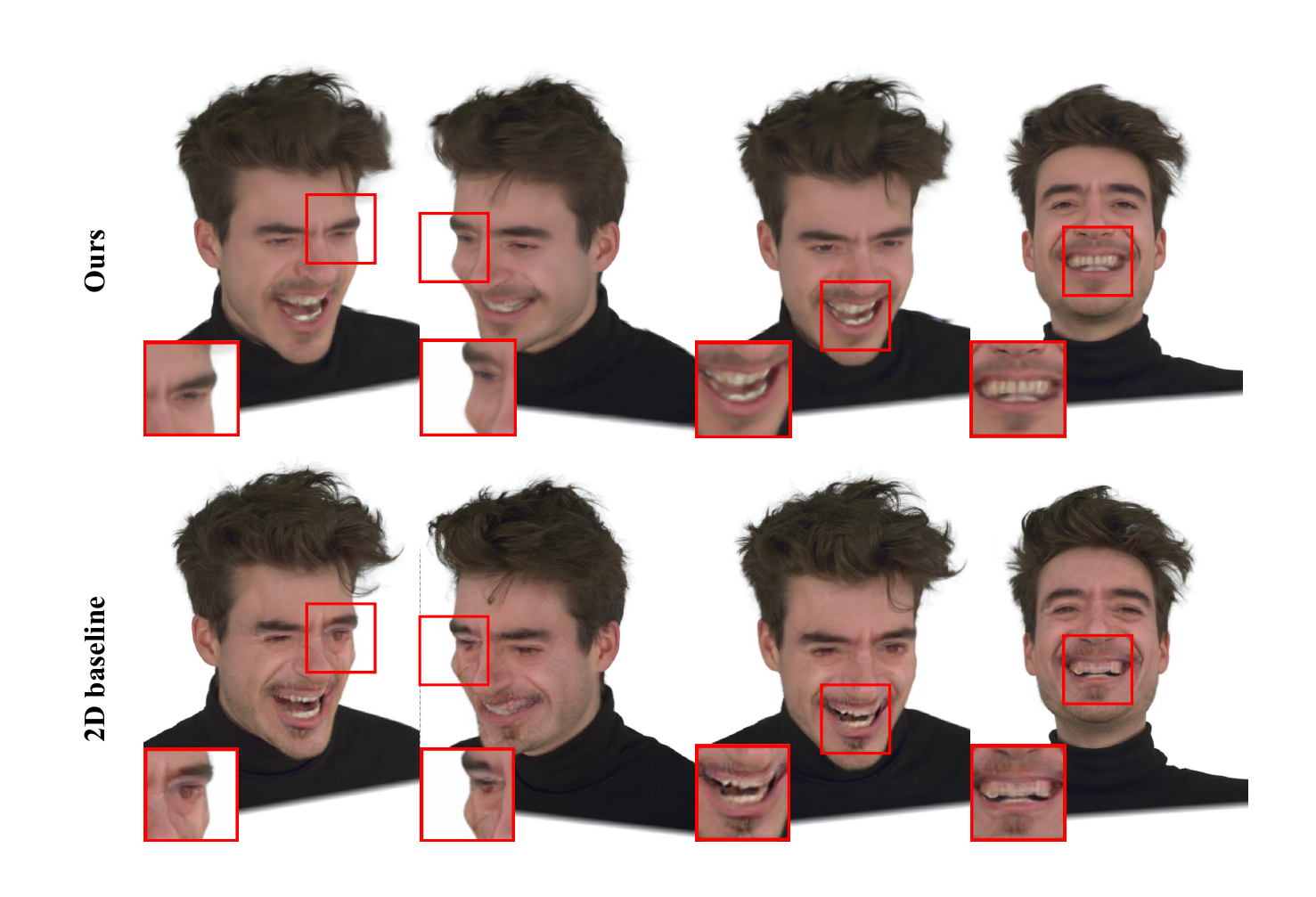} 
    \caption{\textbf{Qualitative Comparison under Extreme Views and Motions.} Top: our method. Bottom: 2D baseline (SimSwap~\cite{chen2020simswap}). Our approach maintains consistent and realistic results across challenging poses and expressions.}
    \label{fig:extreme_view}
\end{figure}

\subsection{Qualitative Comparison with 2D Methods} As shown in Figure~\ref{fig:compare}, our method achieves more photorealistic face swapping results compared to the 2D baseline, particularly under challenging conditions such as exaggerated expressions and extreme head poses. We further evaluate assess the robustness of our method against SimSwap~\cite{chen2020simswap}, as shown in Figure~\ref{fig:extreme_view}. The first row presents our results, while the second row shows those from SimSwap, with zoomed-in facial regions (e.g., eyes and mouth) for clarity.

While SimSwap~\cite{chen2020simswap} performs well on frontal views and neutral expression, its quality degrades under large pose variations or extreme expressions, often producing artifacts such as distorted geometry or texture inconsistencies. In contrast, our method maintains coherent facial structure and expression fidelity across all conditions, benefiting from the strong 3D priors provided by the FLAME model~\cite{li2017flame}. This enables stable identity transfer and structure-preserving synthesis, even in difficult scenarios.

\subsection{Comparison on real-time Facial Reenactment}
\label{sec:fps_comp}

% The results presented in Section~\ref{sec:2d_comp} demonstrate that our method achieves performance comparable to conventional 2D approaches. Notably, thanks to the explicit head geometry provided by 3DGS, our method naturally excels at rendering depth maps with multi-view consistency. Compared to traditional 2D Deepfake methods, our approach poses a greater threat to 3D face recognition systems that rely on depth information for verification.

While existing Deepfake methods can already produce highly realistic face-swapped images, a more alarming application lies in real-time video scenarios, such as video conferencing. These applications require real-time, cross-identity facial reenactment, where facial expressions and head poses are manipulated frame-by-frame.

We compare representative 2D Deepfake methods in terms of runtime efficiency in Table~\ref{tab:fps}. GAN-based approaches offer reasonable speed, while diffusion-based methods are significantly slower due to their iterative nature. Notably, 3DSwap requires test-time finetuning, taking over 3 minutes per image, making real-time use impractical.

In contrast, our 3DGS-based model, once trained, can leverage a real-time FLAME tracker like DECA~\cite{feng2021deca} to predict expressions and render on the fly. Our method achieves 14.45 FPS, about 4× faster than conventional pipelines, fully supporting real-time reenactment.

This underscores a growing security concern: 3DGS-based Deepfake generation makes high-quality, real-time video manipulation not only feasible but efficient, increasing the risk of misuse.

\subsection{Ablation}
In our framework, we introduce a Refine Unit to enhance the quality of the supervisory signals used during training. Since 2D baseline methods like SimSwap may produce artifacts under challenging poses or expressions, directly using their outputs for supervision can degrade 3D results.

To address this, we incorporate a CodeFormer-based~\cite{zhou2022robustblindfacerestoration} refinement module that improves the visual quality of baseline outputs before supervision. This helps correct artifacts in regions prone to occlusion or distortion, such as the mouth and cheeks.

\begin{figure}[htpb]
    \centering 
    \includegraphics[width=\linewidth]{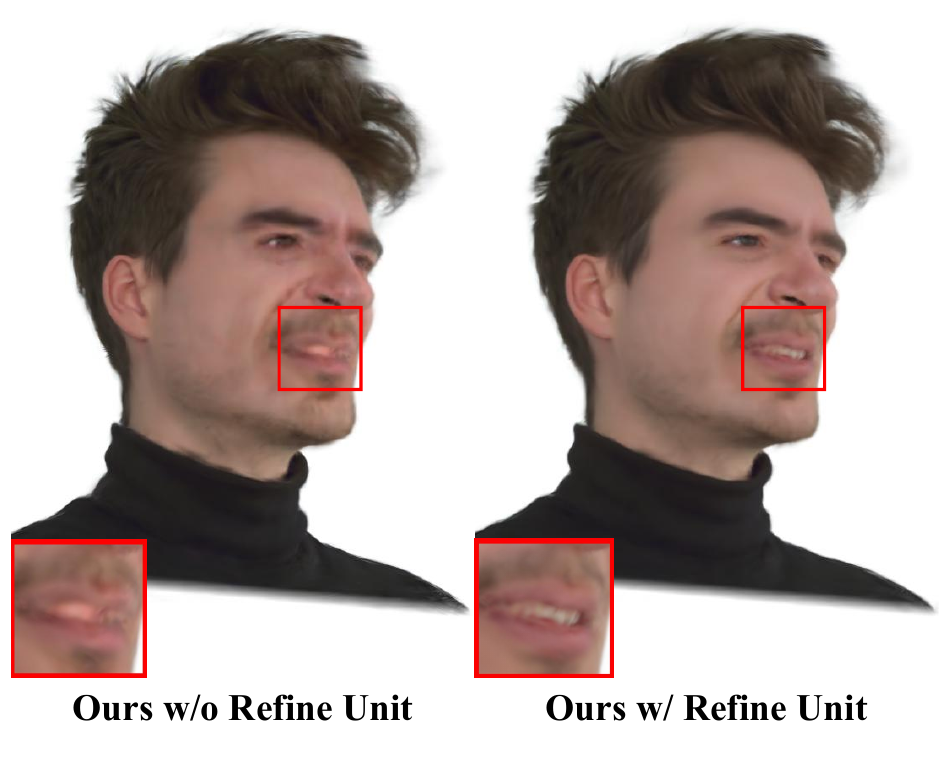} 
    \caption{\textbf{Ablation study on the Refine Unit.} Left: result without the Refine Unit, showing artifacts around the mouth. Right: with the Refine Unit, producing cleaner and more realistic result.}
    \label{fig:ablation}
\end{figure}

As shown in Figure~\ref{fig:ablation}, without the Refine Unit (left), noticeable artifacts appear around the mouth due to noisy supervision. With the Refine Unit (right), these regions are significantly cleaner and more structurally coherent, resulting in more realistic 3D renderings.

\section{Ethical Discussion}
Our experiments not only demonstrate the high visual fidelity of the generated results, but more importantly, reveal the robustness of our method against detection techniques that leverage depth and geometric consistency. Specifically, we show that our 3D-generated outputs remain realistic and coherent even under scrutiny from both 2D and 3D deepfake detectors.

While numerous detection methods have been developed for 2D deepfakes~\cite{sharma2024systematic}, many rely on cues like depth estimation, multi-view consistency, or 3D reconstruction to identify forgeries. These are effective for 2D methods, which often lack true 3D structure. However, little work has addressed detection strategies tailored to 3D deepfake frameworks.

Notably, 3D Gaussian Splatting poses new security risks due to its ability to produce photorealistic, view-consistent, and geometrically coherent outputs, making it harder to detect using conventional approaches. Our method also supports real-time rendering and expression control, making it suitable for scenarios like live video conferencing. This raises serious ethical and security concerns, as such results can be more deceptive than traditional 2D deepfakes.

These findings underscore the urgent need for developing detection strategies specifically for 3D-based forgery methods, and for broader discussion on the ethical implications of controllable, high-quality, real-time 3D deepfake technologies in the age of advanced generative AI~\cite{nguyen2024navigating}.

\section{Conclusion}

In this paper, we proposed a novel 3D Deepfake generation framework based on 3D Gaussian Splatting, enabling realistic, controllable face swapping and reenactment in a fully 3D scene. By combining FLAME-based parametric modeling with point-based rendering, our method supports high-quality identity-preserving synthesis under arbitrary viewpoints. We also introduced a background-aware design and a refinement module to enhance visual consistency under challenging conditions.

Experiments on NeRSemble and physical evaluation show that our method has competitive results on 2D metrics, and excels in 3D consistency and rendering efficiency.

Our work reveals the growing potential of 3D Deepfakes, especially in achieving geometric consistency and real-time performance. We hope it brings attention to this emerging field and encourages further research in both generation and detection.

\section{Acknowledgement}
This work was partially supported by JSPS KAKENHI Grants JP21H04907 and JP24H00732, by JST CREST Grant JPMJCR20D3 including AIP challenge program, by JST AIP Acceleration Grant JPMJCR24U3, and by JST K Program Grant JPMJKP24C2 Japan.

{\small
\bibliographystyle{unsrt}
\bibliography{egbib}
}

% \linenumbers
%\input{X_suppl.tex}

\end{document}